\documentclass[sigconf]{acmart}

\usepackage{array}
\usepackage{stfloats}
\usepackage{url}
\usepackage{verbatim}
\usepackage{bbding}
\usepackage{graphicx}
\usepackage{textcomp}
\usepackage{xcolor}
\usepackage{multirow}
\usepackage{booktabs}
\usepackage{makecell}
\usepackage{pifont}
\usepackage[ruled]{algorithm2e}
\usepackage{arydshln}
\usepackage{tcolorbox}
\usepackage{colortbl}
\usepackage{pifont}

\newcommand{\ie}{\textit{i.e.}}
\newcommand{\our}{Found\xspace} 

\newsavebox\CBox

\definecolor{dark2green}{rgb}{0.1, 0.65, 0.3}
\definecolor{dark2orange}{rgb}{0.9, 0.4, 0.}
\definecolor{dark2purple}{rgb}{0.4, 0.4, 0.8}

\DeclareMathOperator*{\argmin}{arg\,min}
\DeclareMathOperator*{\argmax}{arg\,max}
\usepackage{dsfont}

\usepackage{hyperref}
\usepackage{tablefootnote}
\usepackage{colortbl}
\definecolor{LightCyan}{rgb}{0.88,1,1}
\definecolor{Gray}{gray}{0.92}

\usepackage{balance}

\AtBeginDocument{%
  }

\copyrightyear{2025}
\acmYear{2025}
\setcopyright{cc}
\setcctype{by}
\acmConference[WWW Companion '25]{Companion Proceedings of the ACM Web
Conference 2025}{April 28-May 2, 2025}{Sydney, NSW, Australia}
\acmBooktitle{Companion Proceedings of the ACM Web Conference 2025 (WWW
Companion '25), April 28-May 2, 2025, Sydney, NSW, Australia}
\acmDOI{10.1145/3701716.3715266}
\acmISBN{979-8-4007-1331-6/2025/04}

\begin{document}

\title{Transferable and Forecastable User Targeting Foundation Model}

\author{Bin Dou}
\authornote{The contributions by Bin Dou and Yun Zhu have been conducted completely during internship at Ant Group.}
\affiliation{%
  \institution{Ant Group}
  \city{Hangzhou}
  \country{China}
}
\additionalaffiliation{%
  \institution{Xi'an Jiaotong University}
  \city{Xi'an}
  \country{China}
}
\email{doubin.dou@antgroup.com}

\author{Baokun Wang}
\authornote{Corresponding author.}
\affiliation{%
  \institution{Ant Group}
  \city{Hangzhou}
  \country{China}
}
\email{yike.wbk@antgroup.com}

\author{Yun Zhu}
\authornotemark[1]
\affiliation{%
  \institution{Ant Group}
  \city{Hangzhou}
  \country{China}
  }
\additionalaffiliation{%
  \institution{Zhejiang University}
  \city{Hangzhou}
  \country{China}
  }
\email{zhuyun_dcd@zju.edu.cn}

\author{Xiaotong Lin}
\affiliation{%
  \institution{Ant Group}
  \city{Hangzhou}
  \country{China}
  }
\email{lxt203095@antgroup.com}

\author{Yike Xu}
\affiliation{%
  \institution{Ant Group}
  \city{Hangzhou}
  \country{China}
  }
\email{xuyike.xyk@antgroup.com}

\author{Xiaorui Huang}
\affiliation{%
  \institution{Ant Group}
  \city{Hangzhou}
  \country{China}
  }
\email{huangxiaorui.hxr@antgroup.com}

\author{Yang Chen}
\affiliation{%
  \institution{Ant Group}
  \city{Hangzhou}
  \country{China}
  }
\email{cy462023@antgroup.com}

\author{Yun Liu}
\affiliation{%
  \institution{Ant Group}
  \city{Hangzhou}
  \country{China}
  }
\email{ly319278@antgroup.com}

\author{Shaoshuai Han}
\affiliation{%
  \institution{Ant Group}
  \city{Hangzhou}
  \country{China}
  }
\email{hanshaoshuairs@163.com}

\author{Yongchao Liu}
\affiliation{%
  \institution{Ant Group}
  \city{Hangzhou}
  \country{China}
}
\email{yongchao.ly@antgroup.com}

\author{Tianyi Zhang}
\affiliation{%
  \institution{Ant Group}
  \city{Shanghai}
  \country{China}
}
\email{zty113091@antgroup.com}

\author{Yu Cheng}
\affiliation{%
  \institution{Ant Group}
  \city{Hangzhou}
  \country{China}
}
\email{cy122623@antgroup.com}

\author{Weiqiang Wang}
\affiliation{%
  \institution{Ant Group}
  \city{Hangzhou}
  \country{China}
}
\email{weiqiang.wwq@antgroup.com}

\author{Chuntao Hong}
\affiliation{%
  \institution{Ant Group}
  \city{Beijing}
  \country{China}
}
\email{chuntao.hct@antgroup.com}


\renewcommand{\shortauthors}{Bin Dou et al.}

\begin{abstract}

User targeting, the process of selecting targeted users from a pool of candidates for non-expert marketers, has garnered substantial attention with the advancements in digital marketing. However, existing user targeting methods encounter two significant challenges: (i) \emph{Poor cross-domain and cross-scenario transferability and generalization}, and (ii) \emph{Insufficient forecastability in real-world applications}. These limitations hinder their applicability across diverse industrial scenarios.  
In this work, we propose \textbf{\our}, an industrial-grade, transferable, and \underline{fo}recastable \underline{u}ser targeting fou\underline{nd}ation model. 
To enhance cross-domain transferability, our framework integrates heterogeneous multi-scenario user data, aligning them with one-sentence targeting demand inputs through contrastive pre-training. For improved forecastability, the text description of each user is derived based on anticipated future behaviors, while user representations are constructed from historical information.  
Experimental results demonstrate that our approach significantly outperforms existing baselines in cross-domain, real-world user targeting scenarios, showcasing the superior capabilities of \our. Moreover, our method has been successfully deployed on the Alipay\footnote{https://global.alipay.com/platform/site/ihome} platform and is widely utilized across various scenarios.

\end{abstract}

\ccsdesc[500]{Information systems~Data mining}
\ccsdesc[500]{Information systems~Collaborative filtering}
\ccsdesc[300]{Information systems~Retrieval models and ranking}

\keywords{User Targeting; User Understanding; Self-supervised Pre-training;
Multi-modal Pretraining
}



\maketitle

\section{Introduction}
Recently, user targeting\cite{survey_targeting,simester2020targeting}, the strategic practice of concentrating on specific customers or users to deliver tailored content, messaging, and product recommendations\cite{recommend_survey}, has garnered significant attention with the advancement of digital marketing\cite{chaffey2019digital}. Precise customer targeting forms the cornerstone of any effective marketing strategy, enabling the identification of optimal target audiences for each message and fostering consistent, personalized experiences across channels for individual customers.

To achieve precise user targeting, some previous studies\cite{expert} have developed domain-specific experts tailored to individual scenarios or learning common user embeddings\cite{one4all} through user behavioral sequence data for e-commerce domain. However, expert methods are constrained by their exclusive focus on single scenarios, which limits their transferability and generalization. Additionally, most of them rely on labeled data to train supplementary classification heads for downstream tasks, posing challenges for application in low-resource settings, such as zero-shot learning.  
More recently, LLM-based methods, such as ARALLM\cite{arallm}, aim to enhance generalization across certain scenarios by leveraging vast repositories of common knowledge. Nevertheless, LLMs face significant challenges in utilizing industrial multi-source data, such as payment records and user behavior patterns, due to their insufficient understanding of long-tail knowledge\cite{sun2023head,bang2023multitask}. This limitation undermines the effectiveness of LLM-based methods in complex scenarios, particularly those involving multi-modal temporal and personal data.  
In conclusion, current methods primarily face two challenges: (i) \emph{Poor transferability and generalization}, and (ii) \emph{insufficient forecastability in complex real-world scenarios}.

Therefore, we propose \our, a transferable and forecastable user targeting foundation model designed to address the aforementioned challenges. To enhance transferability and generalization, we develop a user modeling framework that integrates heterogeneous multi-source data, such as billing records, super position model~\footnote{Dataflow statistics generated from page burial} (SPM) logs, mini-program descriptions, user information tables (containing user consumption records, financial irregularities, etc), and search texts within the Alipay application. To improve the forecastability, we collect and curate forecastable user-text pair data to align diverse modalities using contrastive learning. Considering the complexity of Alipay's data, this process is executed in two stages to ensure training stability.  
First, we design two novel self-supervised tasks to pretrain user models, facilitating the extraction of universal embeddings from heterogeneous, multi-scenario user data. Subsequently, we perform user-text alignment, where features from multi-modal data are fused and aligned with representations generated by a fine-tuned LLM through contrastive learning.  
Building on this foundation model, we propose two strategies for utilizing the pretrained models in user targeting: (1) Zero-shot Transfer, which directly inputting requirements in text format to identify targeted users based on user-text similarity. (2) Few-shot Targeting via Prompt-tuning, which Leverages a small set of seed users to enhance the text prompt's descriptive capabilities, thereby improving the performance of text-based user targeting.  
Extensive experiments demonstrate that \our consistently outperforms baseline methods in user targeting, particularly in real-world Alipay applications, where it showcases strong predictive capabilities in complex scenarios. Furthermore, \our achieves superior performance across diverse scenarios and domains, including security, marketing, and recommendation, highlighting its exceptional transferability and generalization.

In summary, our work makes the following contributions:

\begin{itemize}
    \item We propose \our, an industrial-grade, transferable, and forecastable user-targeting foundation model capable of understanding users through large-scale heterogeneous multi-source data to accurately identify targeted users.
    \item We design novel self-supervised tasks for pretraining user models and align different modalities through constructed user-text pair data in subsequent phases, ensuring training stability and forecastability within complex real-world scenarios while enabling the selection of targeted users based on one-sentence demands.
    \item Extensive experiments reveal that our model surpasses baseline methods across diverse scenarios and multiple domains, demonstrating its superior transferability and forecastability in real-world Alipay applications.
\end{itemize}

\section{Related Work}

\subsection{Multi-modal Pre-training}

Multimodal pretraining\cite{multimodal} has become essential for integrating diverse data types, such as text, images, audio, and video, to enhance model performance across various tasks. For instance, CLIP\cite{clip} employs contrastive learning to align image and text embeddings in a shared latent space, enabling zero-shot transfer without task-specific fine-tuning. ALIGN\cite{align} expands this approach by utilizing larger datasets and more complex architectures, achieving state-of-the-art results in image-text retrieval and other benchmarks. BLIP\cite{blip} advances the field by combining contrastive and generative objectives within a unified transformer framework, enriching multimodal representations and improving tasks like image captioning. Additionally, BLIP-2\cite{blip2} and LLaVA\cite{llava} integrate multimodal encoders with large language models, creating robust multimodal large language models. Collectively, these methodologies facilitate the seamless integration of diverse modalities and leverage intermodal connections to boost model performance.

Building upon these advancements, applications in other domains have garnered significant attention, particularly in recommendation systems, by incorporating user-item interactions with multimodal features. This synthesis enhances recommendation accuracy by capturing intricate relationships, as exemplified by models such as MGCL\cite{mgcl}, MMSSL\cite{mmssl}, and MMCPR\cite{mmcpr}. In the graph domain, models such as ConGrat\cite{congrat}, G2P2\cite{g2p2}, and GraphCLIP\cite{graphclip} extend CLIP into graph tasks, with the goal of enhancing zero-shot learning capabilities for graph-related applications.


Unlike existing approaches, we integrate multi-source data of greater complexity. The distinct information across modalities complicates direct model alignment. To overcome these challenges, we propose a two-stage pretraining process: first, pretrain multimodal models using proposed novel auxiliary tasks, and second, align the pretrained models through meticulously constructed data pairs.

\subsection{User Modeling and Targeting}

User modeling focuses on learning transferable user patterns from user behavioral sequence data\cite{msdp,one4all}, while user targeting involves selecting targeted users based on the trained user models. In the context of user modeling, deep representation learning frameworks\cite{ni2018perceive,yuan2020parameter} aimed at obtaining universal user representations are still in their early stages.
For instance, \cite{ni2018perceive} performs multitask representation learning using an attention-based RNN architecture to capture in-depth representations of portal users. 
\cite{yuan2020parameter} propose parameter-efficient transfer learning architecture named PeterRec to relieve the computational cost burden of fine-tuning.
PeterRec covers five downstream tasks that include predicting user profiles like gender, age, and life status, as well as a cold-start recommendation task for browser recommender systems. \cite{zhang2020general} trains autoencoder-coupled Transformer networks that model retention, installation, and uninstallation collectively. 
They test the user embeddings in three downstream tasks for mobile app management scenarios. \cite{gu2021exploiting} proposes a behavioral consistency loss to preserve the user’s longterm interest and an aggregation scheme for the benefit of model capacity. 
The proposed method is evaluated on user preference prediction and user profiling tasks. 

\section{Preliminaries}
In this section, we begin by introducing the notation used in this study (Section~\ref{sec:notations}), followed by an overview of the key downstream tasks addressed in our work (Section~\ref{sec:targeting}).
\subsection{Notations}\label{sec:notations}
In the Alipay application, extensive non-sensitive user information and interactions are accessible. For instance, user behavioral sequence data is denoted as $\mathcal{V} = \{\{B_n\}_{n=1}^N, \{S_n\}_{n=1}^N, \{M_n\}_{n=1}^N\}$, comprising PayBill $B$, Super Position Model $S$, and MiniProgram $M$. Here, $ B_n \in \mathcal{T}^{L^{(B)}_n} $, \( S_n \in \mathcal{T}^{L^{(S)}_n} \), and \( M_n \in \mathcal{T}^{L^{(M)}_n} \) represent the raw text for PayBill information, Super Position Model details, and MiniProgram descriptions, respectively. Let $N$ denotes the number of users and \( n \in [1, 2, \ldots, N] \), \(\mathcal{T}\) is the token dictionary, and \( L_n^{(M)} \) indicates the sequence length.  
Tabular data is denoted as \( T \in \mathbb{R}^{N \times F \times D} \), where \( F \) represents the number of features and \( D \) denotes the dimensionality of each feature. Search text data is represented as \( R \in \mathcal{T}^{L^{(R)}_n} \). This work focuses on the user targeting task, utilizing the aforementioned data to pretrain a user model capable of selecting targeted users based on a single-sentence demand $Q \in \mathcal{T}^{L^{(Q)}}$.


\subsection{User Targeting}\label{sec:targeting}

To develop a user targeting foundation model, extensive user data can be utilized to pre-train a general model endowed with transferable knowledge:

\begin{equation}
\begin{aligned}
g_{\theta^\star}=\argmin\underset{V_n^{\text{s}} \in \mathcal{V}^{\text{s}}}{\mathbb{E}}\mathcal{L}_{\text{pretrain}} & \left(g_\theta;V_n^{\text{s}}\right),
\end{aligned}
\label{equ:pre}
\end{equation}
where $\mathcal{V}^{\mathrm{s}}$ represents the source data of user behavioral sequences, $g_{\theta^\star}$ denotes the pretrained user targeting foundation model, and $\mathcal{L}_{\text{pretrain}}$ refers to our proposed pretraining tasks, which will be detailed in Sec.~\ref{sec:method}.

With the user targeting foundation model trained, the zero-shot transfer for user targeting and few-shot user targeting via prompt-tuning are proposed.
For zero-shot setting, the pre-trained user model can be directly deployed on target data:

\begin{equation}
     p_i = \argmax_{U_i\in\mathcal{U}}P_{\theta^\star}(U_i \mid Q_i)
\end{equation}
where $Q_i$ denotes a single-sentence demand and $U_i$ are the targeted users which require the demand which are sampled from user candidates $\mathcal{U}$.

For the few-shot setting, a limited number of training samples for each class are used for fine-tuning:
\begin{equation}
     g_{\theta^\prime} \in \argmax_{\theta}\mathbb{E}_{(Q_i,U_i) \in \mathcal{I}^{\text{t | tr}}}P_{\theta}(\hat{U}_i = U_i \mid Q_i),
\end{equation}
where $\mathcal{I}^{\text{t | tr}}$ represents the supervised training paired data for downstream user targeting, comprising user demand and the corresponding targeted user group. $\hat{U}_i$ is the predicted user group based on demand sentence $Q_i$. $f_{\theta^\prime}$ denotes the fine-tuned model, which will be evaluated on test samples.


\section{Method}\label{sec:method}
\begin{figure*}[!h]
    \centering
    \setlength{\abovecaptionskip}{0.1cm}
    \includegraphics[width=1.0\linewidth]{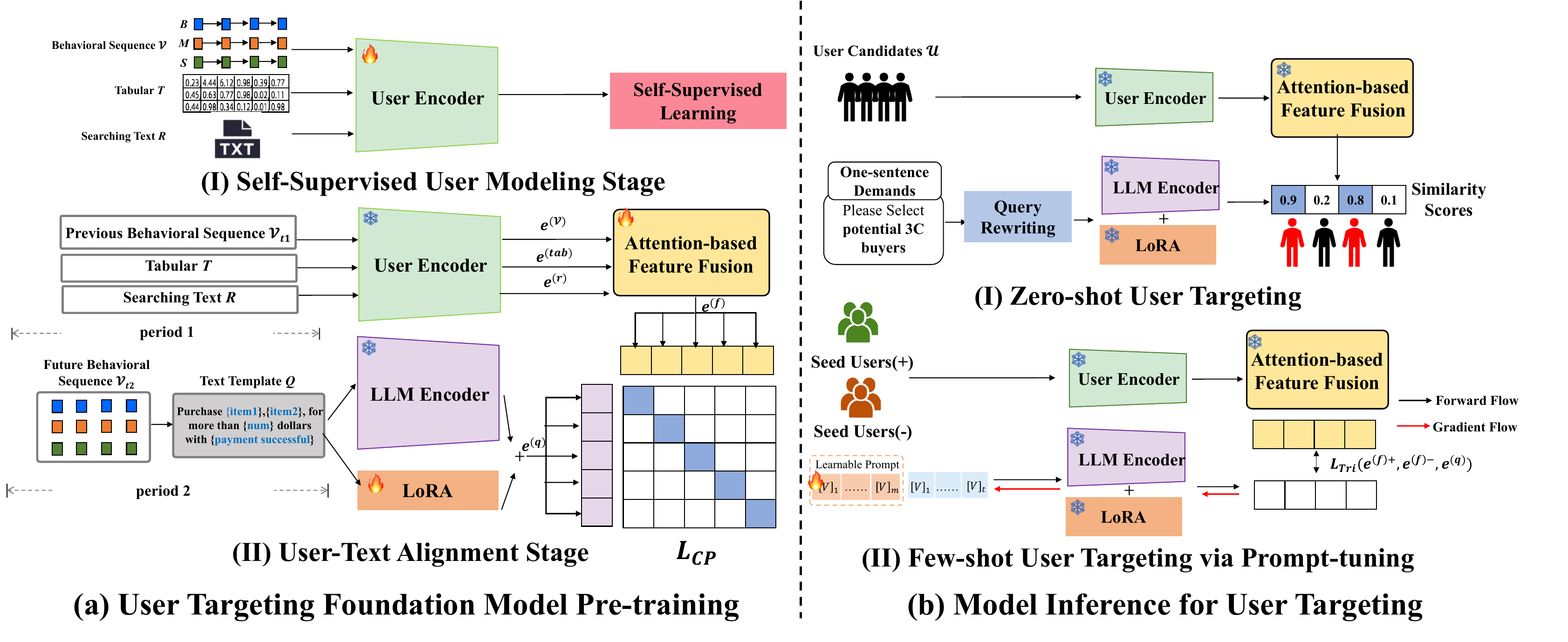}
    \caption{\textbf{Pre-training and inference for user targeting foundation model}. 
        We propose an industrial framework for language-based user targeting, i.e., selecting users from only one sentence.
        it' pre-trained in a two strategy, as the first stage focuses on the self-supervised user modeling while the second is employed to align user-text pairs.
        For user targeting, we develop two approaches: zero-shot transfer, which selects users with one-sentence input, and few-shot user targeting via prompt-tuning.
        }
    \label{fig:model}
\end{figure*}

In this section, we first present the training corpus we collected and curated in Sec.~\ref{user_pair}. Next, we detail the proposed self-supervised tasks for user modeling in Sec.~\ref{sec:ssl}. Subsequently, we explain the user-text alignment process in Sec.~\ref{sec:align}. Finally, we demonstrate the application of our pre-trained model to downstream tasks in Sec.~\ref{sec:down}. The overall pipeline of \our is illustrated in Fig.~\ref{fig:model}.


\subsection{Forecastable User-Text Pair Preparation}
\label{user_pair}

Based on the full amount of user bill, page clicking behavior, miniprogram browsing and ordering, homepage search and other data within the Alipay system, a natural language description of user behavior is constructed. 
It should be noticed that the each user's language description is built based on the future behaviors mentioned before, which will enhance the forecastability of the model pre-trained by the data.
Specifically, each user-text pair can be denoted as:
\begin{equation}
    \mathcal{A}=\{V_{i,t_1}, T_{i,t_1}, R_{i,t_1}; Q_{i,t_2}\}_{i=1}^{N}
\end{equation}
where $t_1$ and $t_2$ denote the previous and future period.

Additionally, different templates are designed according to different data sources, and there are five types in total, divided into purchase category, browsing category, search category, click category and payment channel category.
For example, the billing information occurred by a user will be strung together into a complete description as a user behavioral alignment label based on being at a certain merchant, purchasing a certain product, belonging to a certain category, transaction amount and payment channel.
The template takes the form as:
\begin{tcolorbox}
\textbf{Template for purchase category}: The user purchased \{\textcolor{blue}{items}\} amounting more than \{\textcolor{blue}{num}\} dollars with \{\textcolor{blue}{status}\} payment.
\end{tcolorbox}


In the above template, the placeholders are represented in blue font: ``\textcolor{blue}{items}'' denotes the purchased items, ``\textcolor{blue}{{{num}}}'' represents the amount spent by the user, and ``\textcolor{blue}{{{status}}}'' indicates whether the bill payment was successful. For instance, a user's purchase description could be expressed as:  
"The user purchased \textcolor{blue}{gold}, \textcolor{blue}{gold bars}, etc., amounting to more than \textcolor{blue}{\$10,000} with \textcolor{blue}{successful} payment."  
Templates for other data sources follow similar structures.

\subsection{Self-Supervised User Modeling}\label{sec:ssl}


\subsubsection{User Behavioral Sequence Encoding}

As user behavior contains abundant information,
which greatly improves the generalization ability of user representation and empowers user modeling in many downstream tasks, behavioral sequence is selected 
for user modeling.
At AliPay platform, PayBill $P$, MiniProgram $M$ and SPM $S$ are the main data sources for user representation, thus they are utilized as the input data for user understanding.
Therefore, representational embedding for user $i$ can formulated as:


\begin{equation}
    e^{(\mathcal{V})}_{i} = \mathcal{P}(g_{\theta}(B_{i}, M_{i}, S_{i}))
\end{equation}
where \(\mathcal{P}\) denotes the average pooling function crossing all time windows along each user behavioral sequence, \(g_\theta\) is a composite function \(g_{\theta} = g_\mu \circ g_\nu\), with \(g_\mu\) representing a time-aware function such as GRU\cite{gru} or self-attention networks\cite{transformers}, 
and \(g_\nu\) serving as a text encoding function, such as ALBERT\cite{albert}. Here, \(z = g_\nu(B_{i}, M_{i}, S_{i})\) and \(c = g_\mu(z)\).

\begin{figure}[!h]
    \centering
    \setlength{\abovecaptionskip}{0.1cm}
    \includegraphics[width=1.0\linewidth]{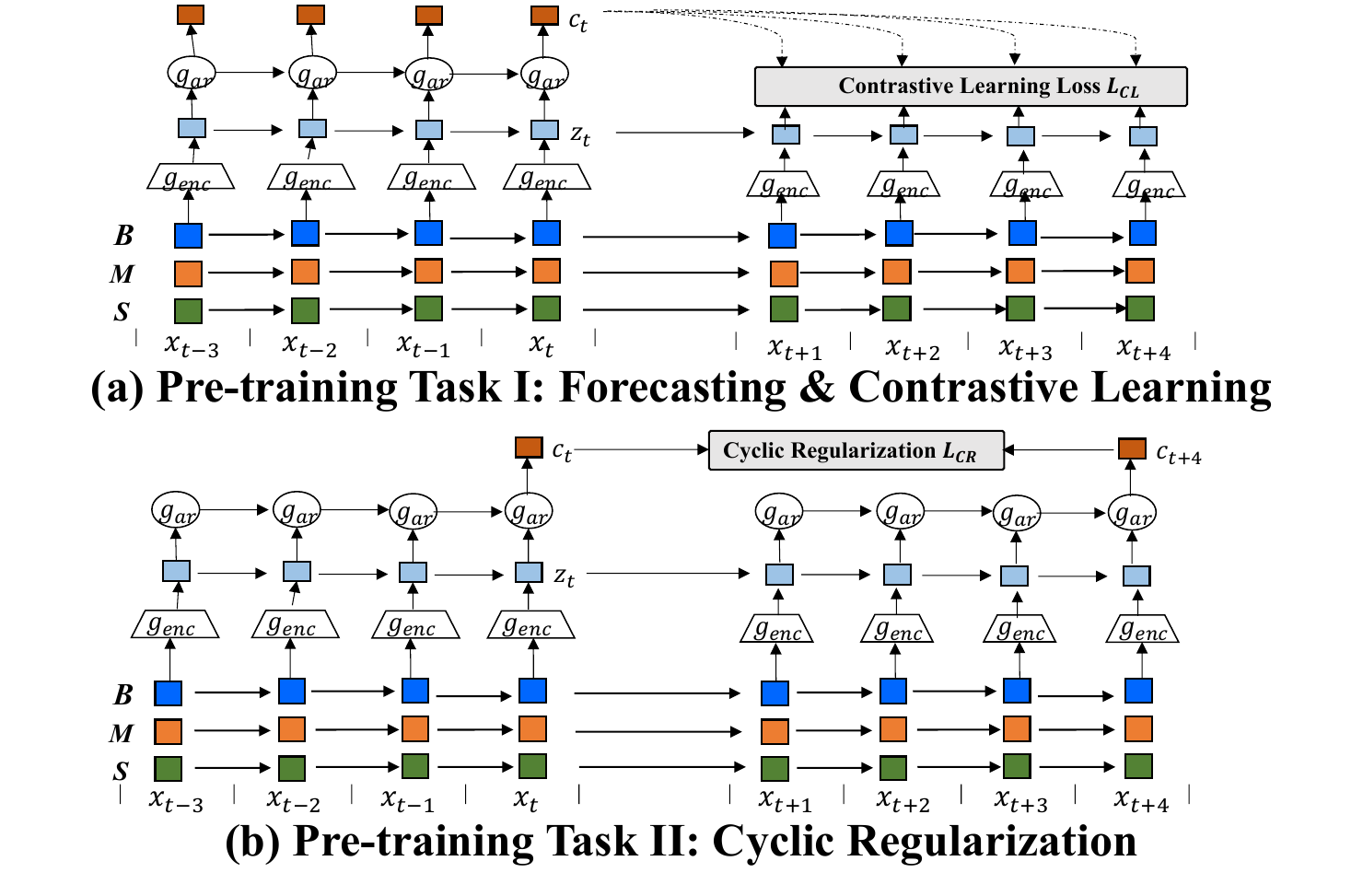}
    \caption{Self-supervised pre-training tasks for user behavioral sequence.}
    \label{fig:ubse}
\end{figure}

As the modals of user  behavioral sequence are represented with text sequences, for User Behavioral Sequence Encoder $g_{\theta}$, we utilize ALBERT\cite{albert}, a light BERT\cite{bert} for self-supervised learning of language representations, to first encode text sequences to embeddings. 
Then, as illustrated in Fig. \ref{fig:ubse} (a), the forecasting prediction is utilized as the main proxy self-supervised pre-training task inspired by Contrastive Predictive Coding\cite{cpc}, as contrastive learning is employed to train the model\footnote{For clarity, we set batch size as 1 here.}:
\begin{equation}
    \mathcal{L}_{\text{CL}}=-\frac{1}{Kk}\sum_{t=1}^{K}\sum_{i=1}^{k}\left[\log\frac{\operatorname{exp}({s(c_{t},z_{t+i}))}}{\sum_{j=1}^{K} \operatorname{exp}({s(c_{t},z_{t+j}))}}\right]
\end{equation}
where \(z_t\) represents the user behavioral embedding at timestamp \(t\) derived from the three data modalities, \(c_t\) denotes the embedding aggregated with temporal information by \(\mu\), \(s\) is the cosine similarity function, \(K\) is the total time length, and \(k \ll K\) controls the window size for positive samples.

Additionally, as the paybill and miniprogram browsing records are commonly presented cyclically, we add cyclic regularization in user behavioral sequence encoding (shown in Fig. \ref{fig:ubse} (b)), enforcing each user's representation to be close across several time windows measured by KL-divergence. Therefore, the total self-supervised pre-training loss for user behavioral sequence is:
\begin{equation}
    \mathcal{L}_{\text{UB}}=\mathcal{L}_{\text{CL}}+\lambda_{cyc} KL(c_t|| c_{t+T})
\end{equation}
where $\lambda$ is the coefficient that controls the strength of the regularization term, $\operatorname{KL}$ denotes the KL divergence between the two input vectors, and $T$ denotes the regularization temporal cycle.

\subsubsection{Tabular Encoding}

In Alipay application scenarios, structural user data can be collected primarily in the form of tabular data. The key features of this tabular data pertain to account security and fund flows, facilitating transferability across multiple domains, such as risk management or marketing, for user targeting.

\sloppy
To obtain the embedding from the user tabular, we employ transformer-based encoder (denoted as $h_{\phi}$).
Specifically, given an input table 
$T = \{ T_{cat}, T_{cont} \}$, where $T_{cat}
$ represents categorical tabular features and $T_{cont}
$ represents continuous features, $h_{\phi}$ processes $T_{cat}$ similar to \cite{tabtransformer} (encoded with Column Embedding \cite{tabtransformer} and fed into Transformer) while handles $T_{cont}$ by mapping it to another feature item with MLP and also using it as the input to the transformer-based encoder.
The corresponding embedding $e^{\text{(tab)}}$ for user $i$ can be denoted as:
\begin{equation}
    e^{\text{(tab)}}_i = h_{\phi}(T_i)
\end{equation}

To pre-train the encoder, 
two strategies are employed: masked language modeling (MLM)\cite{bert} and replaced token detection (RTD)\cite{li2022pre}. 
For MLM,
we randomly mask part of the categorical features $\tilde{T}_\mathcal{M} \subset T_{cat}$. The encoder is trained by minimizing the cross-entropy loss of a multi-class classifier that predicts the original values of the masked features based on the Transformer generated embeddings:
\begin{equation}
    \mathcal{L}_{\text{MLM}}=\sum_{m\in\mathcal{M}}\operatorname{CE}(\hat{\tilde{T}}_m, \tilde{T}_m)
\end{equation}
where $\mathcal{M}$ represents the set of masked feature IDs, $\operatorname{CE}$ is the cross entropy loss, $\hat{\tilde{T}}_m$ and  $\tilde{T}_m$ denote the prediction of masked feature and the ground truth respectively.
In the RTD approach, the original categorical feature values are replaced with random values from the same feature set. A binary classifier is then trained to predict whether a feature has been replaced, with the loss minimized accordingly, which can be expressed as:
\begin{equation}
    \mathcal{L}_{\text{RTD}} = \sum_{r\in\mathcal{R}}\operatorname{BCE}(\hat{\tilde{T}}_r, \tilde{T}_r)
\end{equation}
where $\mathcal{R}$ represents the replaced features, $\operatorname{BCE}$ is the binary classification loss, $\hat{\tilde{T}}_r$ is the prediction, and $\tilde{T}_r \in \{ 0, 1 \}$ denotes whether the $r$-th feature is replaced. Based on the aforementioned tasks and hyperparameters $\lambda_{\text{MLM}}$ and $\lambda_{\text{RTD}}$, the tabular pre-training loss is:
\begin{equation}
    \mathcal{L}_{\text{Tab}}=\lambda_{MLM}\mathcal{L}_{\text{MLM}}+\lambda_{\text{RTD}}\mathcal{L}_{{\text{RTD}}}
\end{equation}

\subsubsection{Text Encoding}

At Alipay platform, user searching text can also assist to achieve user understanding.
Therefore, we utilize a light BERT model, \ie, AlBERT\cite{albert} as the encoder $f_{\omega}$, to encode the user $i$'s searching text $R$, which is pre-trained similar to \cite{li2022pre}. The searching text encoding can be expressed as:
\begin{equation}
    e^{\text{(r)}}_i = f_{\omega}(R_i)
\end{equation}

\subsection{User-Text Alignment}\label{sec:align}

\subsubsection{Attention-based User Feature Fusion}

For the multi-modal user representation, the embeddings from different sources are fused using cross-attention \cite{crossattn}, which is a common feature aggregation approach for multi-modal features. The fused embedding of the user $i$ can be expressed as :
\begin{equation}
    e^{f}_{i} = \operatorname{CA}(e^{(\mathcal{V})}_{i}, e^{({\text{tab}})}_{i}, e^{(\text{{r}})}_{i})
\end{equation}
where $\operatorname{CA}$ represents the cross-attention among all data modalities.

\subsubsection{Contrastive User-Text Pre-training}

To achieve user targeting task  with  only one sentence input, we develop the user targeting foundation model inspired by CLIP\cite{clip}, which aligns multi-modal features with contrastive learning.
Specifically, each user's description text is developed according to Sec. \ref{user_pair}, which are fed into the LLM encoder fine-tuned with LoRA\cite{lora} to generate the user text embedding:

\begin{equation}
    e^{q}_{i} = \operatorname{LLM}(\mathcal{F}(B_{i,t_2}, M_{i,t_2}, S_{i,t_2}))
\end{equation}
where $\mathcal{F}$ represents the process of generating user description according to the template illustrated in Sec. \ref{user_pair}.

After obtaining the user and the corresponding encoding of behavior sequence text, we employ contrastive loss Info-NCE \cite{cpc} to align the modalities, \ie, fused user embedding $e^f_{i,t1}$ generated from previous multi-modal user data and text embedding $e^q_{i,t2}$. The loss is applied to pulling together the representation from positive pairs while pushing apart negative pairs, takes the form as:

\begin{equation}
    \mathcal{L}_{\text{CP}}=-\frac{1}{B}\sum_{i=1}^{B}\left[\log\frac{\operatorname{exp}({s(e^f_{i,t_1},e^q_{i,t_2}))}}{\sum_{j=1}^{B} \operatorname{exp}({s(e^f_{i,t_1},e^q_{j,t_2}))}}\right]
\end{equation}
where $B$ means the batch size and $s$ denotes the cosine similarity function.

\subsection{User Targeting}\label{sec:down}

In this section, we introduce the techniques employed to adapt models for user targeting. 
Figure \ref{fig:model} (b, I) illustrates the main user targeting pipeline of our model. 
First, we illustrate the adaptation of our model on target data for zero-shot transfer. Then, we propose a novel prompt tuning method for few-shot user targeting.

\subsubsection{Zero-shot Transfer}

Once pre-trained, our model can be directly deployed on target datasets without any additional training, i.e., enabling zero-shot inference according to the one-sentence input as depicted in Figure \ref{fig:model} (b, I). 

Additionally, to enable our model accessible to language demand from non-experts, we add a module to rewrite the demand, matching it with the template designed during pre-training and enhancing its cognitive ability for user data inside Alipay application. 
Therefore, the module performs as the \textbf{Query Rewriting}. We design a simple two-stage fine-tuning scheme: the first stage is designed to refer to scaling-law to synthesize a multi-round conversation set using multi-source data from Alipay to inject domain knowledge to one LLM (QWEN2-14B\cite{qwen2}); then the seconda stage is to mine different people's circling intentions under <Query - Answers> for fine-tuning.


\subsubsection{Few-shot User Targeting via Prompt Tuning}

In low-resource scenarios, where only a few user samples (named seed users) exist for the targeting task, we also design an approach for few-shot learning via prompt-tuning, as illustrated in Figure \ref{fig:model} (b, II).

Inspired by previous studies\cite{coop, cocoop, prograd}, the seed users can be utilized as labels to learn contexts which improves the descriptive ability of the prompt, by adding learnable tokens to the input text.
In practical user targeting scenarios, some negative samples share the similar behaviors with the ground-truth samples. 
For example, 3C digital repairers may have the similar behaviors compared to 3C digital enthusiasts.
They can be defined as the hard-negative samples in user targeting task and the mistargeting of them may cause risk in real-world scenarios. 
To handle the hard-negative samples in our few-shot learning process, we introduce the negative samples $U^-$, which can obtained from business feedback, into prompt-tuning together with positive seed users $U^+$.
Through the contrastive loss, \ie, triplet loss\cite{triplet}, the learned prompt embedding.
\begin{equation}
    \mathcal{L}_{\text{Tri}}(e^{f+}, e^{f-}, e^{q}) = \sum \operatorname{max}(0,\|e^{q}-e^{f+}\|^{2}-\|e^{q}-e^{f-}\|^{2}+\alpha)
\end{equation}
where $e^{f+}$, $e^{f-}$ and $e^{q}$ denote the embeddings of positive seed users $U^+$, negative seed user $U^-$ and the learnable text embedding, respectively, $\alpha$ is the threshold to control embedding similarity. 

Once trained, the tuned prompt can be used as the input to achieve user targeting via pipeline in Figure \ref{fig:model} (b, I).


\section{Experiments}

In this section, we begins with the experimental setup, detailing the datasets, baselines, evaluation metrics, and implementation specifics (Section~\ref{exp:setup}). Next, we present extensive experiments that demonstrate our model’s effectiveness in user targeting and representation (Section~\ref{exp:main}). Finally, we perform an ablation study to evaluate the effectiveness of each proposed component (Section~\ref{exp:ablation}).

\subsection{Experimental Setups}\label{exp:setup}

\subsubsection{Datasets}

For pre-training the industrial framework of
user targeting, we employ the real-world industrial dataset, which collects heterogeneous multi-scenario user data on Alipay. 
Specifically, the training dataset $\mathcal{D}_{train}$ contains around 500 million pieces of user information (user-text pairs), with each piece organized according to Section \ref{user_pair}.
The user targeting test benchmark contains 13 scenarios, covering security risk control, recommendation and marketing domain.

The dataset of each scenario contains 0.3-30 million pieces of user information, which are in the format $\mathcal{D}_{test}=\{(u_i, l_i)\}_{i=1}^{N}$,
as $u_i$ represents the same user data modal as in training while
label $l_i$ ranges from $\{ 0,1 \}$, denoting whether the user conducts the operations in the domain, which is annotated according to real-world business scenario.
For linear probe analysis, each test dataset is divided 1:4, of which the first part is used for linear layer fitting (named \textit{fitting part}) while the second (named \textit{evaluation part}) is used for evaluation.
Table \ref{tab:dataset} shows the statistics of our training and test dataset.

\begin{table}[htp]
    \caption{Data statistics for user foundation model pre-training as well as the test benchmarks.}
    \centering
    \resizebox{\linewidth}{!}{
    \begin{tabular}{c|c|cc|c}
    \toprule
\textbf{Dataset}  & \textbf{No.} & \textbf{Domain} &  \textbf{Scenario} &  \textbf{Number(d)}
 \\ \midrule
$\mathcal{D}_{train}$ & -- &  General       &    General     &   $\approx 500,000,000$       \\ \midrule
 & \#1 & Account Security      &     Account Theft Risk      &     2,910,925      \\
 & \#2 &  Marketing       &     APP Click     &    334,769       \\
$\mathcal{D}_{test}$ & \#3 & Marketing  &   APP Activity        &   743,828              \\
 &  \#4 &  Marketing       &     Willingness for Phone Renting     &     280,302      \\
 &  \#5 &  Recommendation       &   Willingness for Credit Facilities       &     6,435,977      \\
\bottomrule
    \end{tabular}}

    \label{tab:dataset}
\end{table}

\subsubsection{Baselines}

The following state-of-the-art user behavioral understanding and targeting models are utilized in our comparison experiments.

\begin{table*}[htp]
    \caption{Quantitative comparison on zero-shot user targeting performance on Alipay benchmarks. Accuracy / Precision / Recall are presented.}
    \centering
    \resizebox{\linewidth}{!}{
    \begin{tabular}{c|ccccc}
    \toprule
Methods  &  $D_{test}\#1$ & $D_{test}\#2$ & $D_{test}\#3$ & $D_{test}\#4$ & $D_{test}\#5$ \\ \midrule

ARALLM(GPT-3.5)  &      77.1/69.6/84.7        &     87.8/86.7/85.0      &    52.5/53.0/56.2      &   56.6/50.8/68.8   &   64.9/53.3/73.0   \\

ARALLM-ft(ChatGLM2-6B)  &     83.8/74.4/91.6     &     94.8/90.7/89.0      &    58.0/54.8/61.0      &    63.6/52.7/71.6    &  78.6/65.7/78.8 \\

ARALLM-ft(QWEN2-7B)  &    83.2/76.0/91.8      &    95.4/91.2/89.0       &    57.0/54.5/61.8      &   64.9/53.3/73.0     &  79.3/68.3/82.1  \\

Ours    &     \textbf{88.6/83.7/94.1}        &     \textbf{96.0/92.3/90.0}      &  \textbf{61.9/57.6/67.0}        &   \textbf{67.6/60.4/76.9}   &  \textbf{84.3/77.3/80.0}   \\

\bottomrule
    \end{tabular}}
    \label{tab:zeroshot}
\end{table*}

\begin{table*}[htp]
    \caption{Quantitative comparison on few-shot user targeting. 10 seed users are utilized for all methods. Accuracy / Precision / Recall of user targeting performance are presented.}
    \centering
    \resizebox{\linewidth}{!}{
    \begin{tabular}{c|c|ccccc}
    \toprule
Model  &  $D_{test}\#1$ & $D_{test}\#2$ & $D_{test}\#3$ & $D_{test}\#4$  
& $D_{test}\#5$ & \\ \midrule

U-MLP One4all  &    91.0/86.8/93.2      &     96.3/94.0/93.1      &     77.0/64.5/81.8     &    76.6/70.7/83.8    &  87.6/72.7/88.8  \\


ARALLM-ft(ChatGLM2-6B)  &    87.8/79.0/90.7     &     95.6/92.8/90.4      &     71.8/62.2/77.0   &    69.1/65.3/78.3   &   89.1/74.1/90.2   \\

ARALLM-ft(QWEN2-7B)  &     88.1/81.6/91.7     &     95.7/93.0/90.8      &   71.0/60.8/76.2       &    71.6/65.7/78.8    &  89.9/78.3/91.0  \\

Ours  &  \textbf{95.6/93.8/97.4}         &     \textbf{97.1/94.6/93.7}      &   \textbf{81.7/71.4/87.1}       &   \textbf{84.3/80.0/88.9}   &  \textbf{93.0/87.7/95.2}  \\
\bottomrule
    \end{tabular}}

    \label{tab:few-shot}
\end{table*}



\begin{itemize}
    \item \textbf{ARALLM}\cite{arallm} 
    Analogical Reasoning Augmented Large Language Models, which is the LLM-based method and achieves user targeting from one-sentence demand by structurally understanding the input sentence and matching the users representation with the labels. 
    Pre-trained as well as Question-Answering fine-tuned models can be used for user targeting.
    Here we employ GPT-3.5\cite{ouyang2022training} as the pre-trained model, open-source LLMs as the base model for fine-tuning, such as ChatGLM\cite{chatglm} and QWEN2\cite{qwen2}. 
    \item \textbf{U-MLP One4all}\cite{one4all}. One4all is a general-purpose representation learning through large-scale pre-training, while U-MLP is one of its extended user targeting model which adds MLP user decoder to generate targeted users.
    \item \textbf{MSDP}\cite{msdp}
    Multi-scale Stochastic Distribution Prediction model for learning user behavioral sequence representation, which takes the prediction on user’s behaviors distribution over a period of time as the self-supervision signal.
\end{itemize}

In addition,\textbf{ TabTransformer trained with RTD \& MLM}\cite{tabtransformer} tasks is employed as the baseline for tabular modality (tabular encoding model), \textbf{QWEN2-7B}\cite{qwen2} is employed as the baseline for text modality (searching text encoding model) for user representation evaluation.

\subsubsection{Evaluation Metrics}

For \textbf{User Targeting Task}, as in each scenario the task can be treated as binary classification, we employ the \textit{Accuracy / Precision / Recall} metrics for evaluation. 
For \textbf{User Representation}, linear probe representation analysis is conducted on all annotated datasets, which is evaluated by \textit{AUC} (Area Under the ROC Curve\cite{auc}) and \textit{KS} (Kolmogorov-Smirnov\cite{berger2014kolmogorov}) metrics.

\subsubsection{Implementation Details}

The embedding sizes of user representation and text features are set to 1024-dim.
In Self-Supervised User Pre-training stage, the user behavioral sequences, i.e., paybill, SPM and miniprogram, which are in the form of text, are first encoded with AlBERT\cite{albert}. Then 6-layer Transformer is used to generate encoding of the series for further pre-training.
For user tabular information, the similar configuration of TabTransformer is utilized in the tabular encoder.
In the following User-Text Alignment stage, embeddings from multi-modal user features are fused using 5 cross attention layers.
To generate the encoding of user text description, QWEN2-1.8B\cite{qwen2} is utilized as the LLM encoder, which are tuned with LoRA\cite{lora} together with user encoding model. 
For both of stages, the optimizer is AdamW, with cosine decay learning rate initialized at 4e-4. 
The model is pre-trained on 16 A100 GPUs (80G), and a single A100 is used for testing (linear probe for user embedding and zero-shot, few-shot user targeting).


\subsection{Comparisons}\label{exp:main}

\subsubsection{Zero-shot User Targeting}
We take user targeting via zero-shot transfer based on one sentence input on the benchmarks. 

\textit{Experimental Setup.}
Both ARALLM and fine-tuned ARALLM are also employed as the baseline, taking the same sentence prompt as input. 

\textit{Analysis.}
Quantitative results in Table \ref{tab:zeroshot} demonstrate the superiority of our model, as it outperforms baselines 10\%/15\%/10\% at Accuracy/Precision/Recall metrics crossing all domains.

\subsubsection{Few-shot User Targeting}
We also take few-shot user targeting on the benchmarks. 

\textit{Experimental Setup.}
Both ARALLM and fine-tuned ARALLM are also employed as the baseline, together with U-MLP One4all model trained with the seed users. 
For all few-shot learning baselines, the number of user samples is kept at 10. To ensure the comparison fairness upon seed user number, our model is trained with 5 positive and 5 negative users (10 total).

\textit{Analysis.}
The quantitative results in Table \ref{tab:few-shot} demonstrate the superiority of our model, as our model outperforms the baselines 5\%/8\%/5\% at Accuracy/Precision/Recall metrics crossing all domains.

\begin{figure}[!h]
    \centering
    \setlength{\abovecaptionskip}{0.1cm}
    \includegraphics[width=1.0\linewidth]{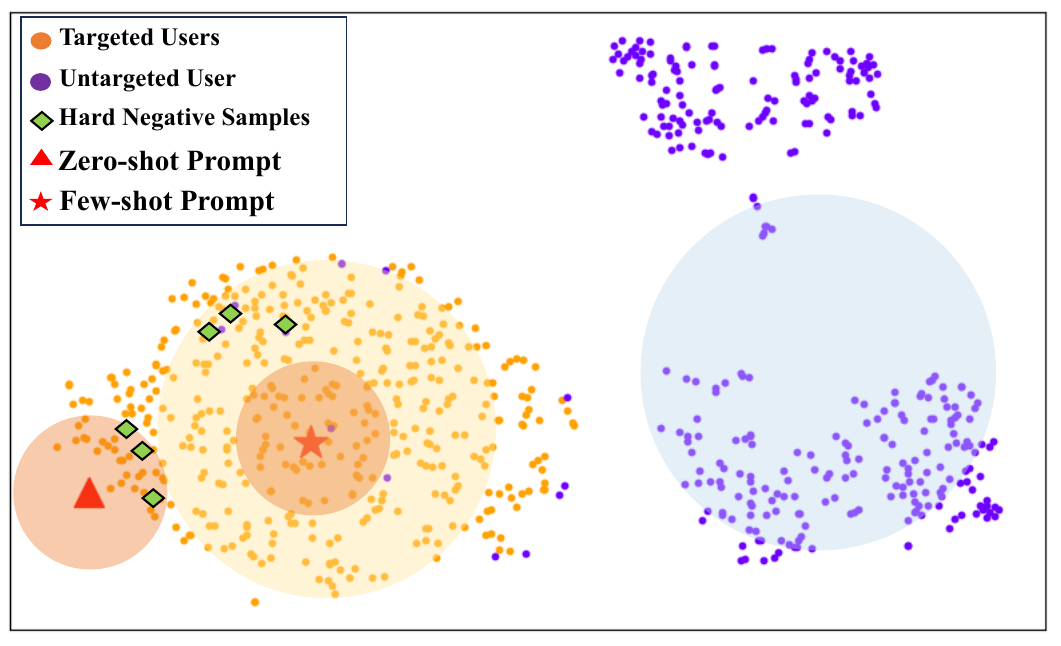}
    \caption{
    Few-shot User Targeting.
    The embeddings of both targeted, untargeted and hard-negative users are visualized with t-SNE, presented in the same coordinate system together with prompt embedding.
    }
    \label{fig:few_shot}
\end{figure}

The visualized results are presented in Figure \ref{fig:few_shot}, which exhibit the performance improvement of the few-shot learned embedding intuitively. Also, the proximity to positive samples and distance from hard-negative samples proves the effectiveness of our approach, especially for handling hard-negative users.

\subsubsection{User Representation Evaluation}

We also evaluate the performance of the pre-trained user representation from both our model and other user modeling baselines.

\textit{Experimental Setup}
For representation evaluation, we utilize linear probe analysis, which means employing linear network layers to the user embeddings to fit the  \textit{fitting part} of the dataset and evaluate on the \textit{evaluation part}.

\begin{figure}[!h]
    \centering
    \setlength{\abovecaptionskip}{0.1cm}
    \includegraphics[width=\linewidth]{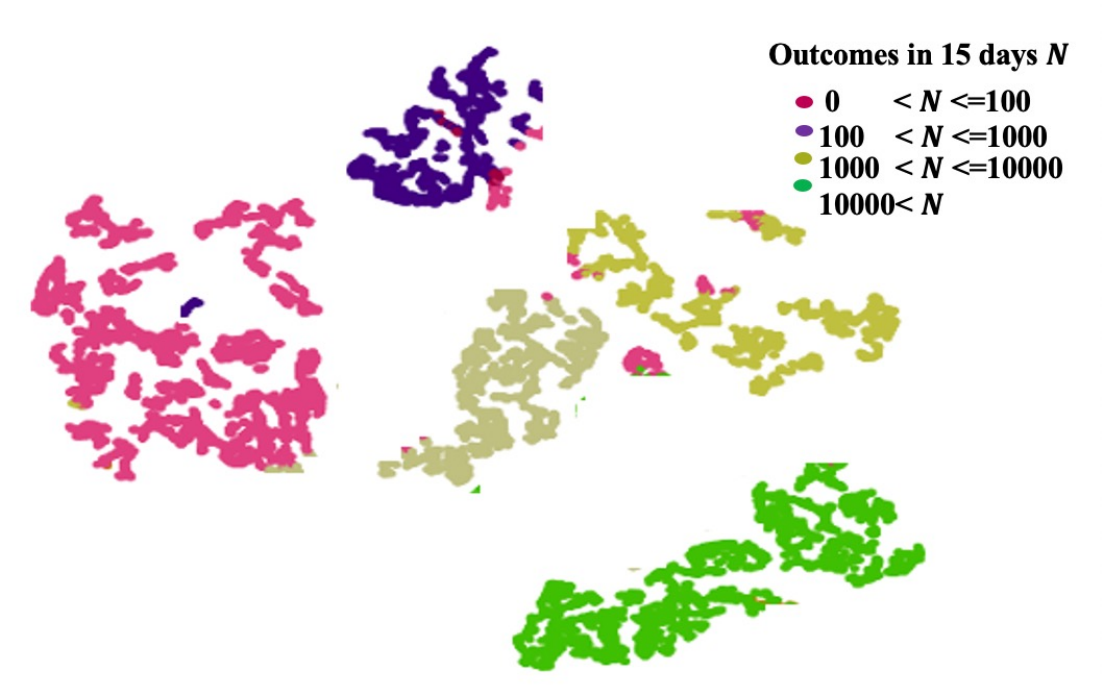}
    \caption{t-SNE visualization of user representations. 
    Every color represents users with different attributes 
    (here denoting the outcomes of each user).
    }
    \label{fig:user_rep}
\end{figure}

\textit{Analysis}
As shown in Table \ref{tab:linear_probe},
the representation from our user model outperforms other baselines trained by single-modal data (+ 4\%/2\% than User Behavioral Sequence, + 12\%/13\% than Tabular data, + 11\%/10\% than Searching Text).
The improvement from simple feature fusion via concatenation also demonstrates the effectiveness of our user feature fusion model (attention-based fusion guided by text description).
Also, Figure \ref{fig:user_rep} shows the visualization of user embeddings, as users from each consuming scenario presented as one color, demonstrating the user representation capabilities of our model.

\subsection{Ablation Studies}\label{exp:ablation}

We conduct ablation studies on user data modal, feature fusion module and the user-text alignment method.

\begin{table*}[htp]
    \caption{Linear probe results. AUC/KS metrics on Alipay user targeting benchmarks are presented.}

    \centering
    \resizebox{\linewidth}{!}{
    \begin{tabular}{c|c|ccccc}
    \toprule
Methods & Data Modal &  $D_{test}\#1$ & $D_{test}\#2$ & $D_{test}\#3$ & $D_{test}\#4$ & $D_{test}\#5$ \\ \midrule

U-MLP One4all & User Behavioral Sequence &  85.5 / 60.2  &  86.8 / 61.7    & 83.3 / 56.4 & 93.3 / 69.6 & 94.6 / 78.9 \\

MSDP & User Behavioral Sequence &  86.4 / 60.6       &    86.7 / 61.2      & 84.6 / 58.0 & 92.5 / 69.5 & 95.7 / 83.3 \\

\hdashline

 TabTransformer-MLM\&RTD & Tabular &  77.9 / 49.9       &    78.5 / 57.3      & 74.6 / 53.0 & 82.5 / 69.9 & 85.8 / 71.1 \\
\hdashline

QWEN2-7B & Text &  78.7 / 52.9       &    78.9 / 57.6      & 72.7 / 44.7  & 88.6 / 61.3 & 88.6 / 73.0 \\
\hdashline

Concatenation & Multi-source Heterogeneous & 86.6 / 61.0 & 87.5 / 62.3 & 85.7 / 60.2 & 93.7 / 70.6 & 96.1 / 86.4
  \\

Ours             & Multi-source Heterogeneous &   \textbf{89.3 / 62.7}       & \textbf{90.7 / 65.9} & \textbf{88.7 / 61.5} & \textbf{95.2 / 73.9} & \textbf{98.9 / 90.7}         \\
\bottomrule

    \end{tabular}}
    \label{tab:linear_probe}
\end{table*}

\begin{table*}[htp]
    \caption{Ablation experiments on zero-shot user targeting over Alipay benchmarks.
    Different data sources and feature fusion module in pre-training are ablated in the experiment.
    Accuracy / Precision / Recall of user targeting are presented.
    }
    \centering
    \resizebox{\linewidth}{!}{
    \begin{tabular}{c|ccccc}
    \toprule
Methods  &  $D_{test}\#1$ & $D_{test}\#2$ & $D_{test}\#3$ & $D_{test}\#4$ & $D_{test}\#5$ \\ \midrule
w/o User Behavioral Sequence   &     82.6/75.5/88.0      &     89.7/83.8/85.8     &   52.6/50.5/56.9   &  58.7/52.9/65.0   &  76.7/65.6/74.7   \\

w/o Tabular  &    84.3/79.3/90.0      &     91.3/88.1/86.5     &     55.0/55.9/63.3     &   63.3/59.4/85.7    &  93.5/90.1/89.7  \\

w/o Search Text    &    83.6/78.7/89.7    &   90.2/87.7/86.4    &   54.4/53.9/62.5    &  60.2/56.4/68.6  &  79.9/71.2/77.8  \\

w/o Attention-based Fusion    &    87.3/81.6/93.5    &   93.7/89.5/88.8    &   58.7/56.3/64.7    &  64.3/58.3/72.7  &  83.9/75.9/79.6 \\

Full    &     \textbf{88.6/83.7/94.1}        &     \textbf{96.0/92.3/90.0}      &  \textbf{61.9/57.6/67.0}        &   \textbf{67.6/60.4/76.9}   &  \textbf{84.3/77.3/80.0}   \\

\bottomrule
    \end{tabular}}

    \label{tab:ablation}
\end{table*}

\subsubsection{Pre-training Data Modal}

To verify the effect of 
the multi-source pre-training data, we conduct the comparison between the model trained without different data modal and the full model.
Comparison between Table \ref{tab:ablation} row 1-3 and row 5 shows the effectiveness of all data sources for pre-training our foundation model, while the user behavioral sequence plays the most significant role in user understanding, leading to over 6\%/8\%/6\% Accuracy/Precison/Recall.

\subsubsection{Attention-based Feature Fusion}

The effect of feature fusion module in our user-text alignment stage is also evaluated.

\textit{Experimental Setup.}
We employ the feature concatenation instead of the attention-based fusion module to fuse multi-modal user features as comparison.

\textit{Analysis.}
As shown in 4-th row of Table \ref{tab:ablation}, the absence of attention-based multi-modal user feature fusion will lead to a performance decrease of $\sim$ 1\%/2\%/1\% Accuracy/Precison/Recall.

\section{Conclusion}

In this paper, we propose an industrial framework \our for transferable and predictive user targeting.
Our model takes one-sentence form demand as input and achieve user understanding with heterogeneous multi-scenario data, enabling it to select targeted user among candidates.
The framework is pre-trained in a two-stage strategy, where the first Self-Supervised User Modeling Stage is built to generate user representation from multi-modal user data, and the second User-Text Alignment Stage is built for fusing multi-modal user features and align user embedding with corresponding language representation for user targeting.
Additionally, to enhance the model predictive ability, the user-text pair for 
our framework pre-training is also designed, as the text is obtained via from future use behaviors while user representation is generated from pervious information.
With the pre-trained framework, zero-shot transfer and few-shot user targeting via prompt-tuning are designed to achieve user targeting.
On real-world benchmarks, our model outperforms baselines in cross-domain scenarios, which demonstrates the superiority of our model.
Additionally, in real-world application, our model exhibits superior performance compared with other industrial frameworks. 
Since 2024, \our is deployed in the user selection platform of Alipay for numerous domains, obtaining an improvement for different user understanding scenarios. For example, improving 30\% CTR for "Used Car Selling Preferences" users and 13.2\% CTR for "Video Watching in Alipay" users.

\begin{acks}
This work was supported by the National Key Research and Development Plan of China (2023YFB4502305), and Ant Group through Ant Research Intern Program.

We acknowledge that the datasets are only used for academic research, and do not represent any real business situation. They are desensitized and encrypted, do not contain any Personal Identifiable Information and were destroyed after the experiments. Adequate data protection was carried out during the experiment to prevent the risk of data copy leakage. 
\end{acks}

\clearpage

\bibliographystyle{ACM-Reference-Format}
\balance
\bibliography{refs}

\clearpage

\appendix

\section{Additional Model Details}

\subsection{Self-Supervised User Modeling}
\label{self-supervise}
For \textbf{User Behavioral Sequence Pre-training}, the time span for each window represents one week. 
The behavioral sequence length is 9, divided into 5 (fed into the time-aware autoregressive model) and 4 (used as the future behaviors for contrastive learning).
$\lambda_{cyc}$ is set at 0.1.
The user behavioral sequence pre-training takes $\sim$ 40 hours on 8 A100 GPUs for model convergence.

\begin{figure}[!h]
    \centering
    \setlength{\abovecaptionskip}{0.1cm}
    \includegraphics[width=0.8\linewidth]{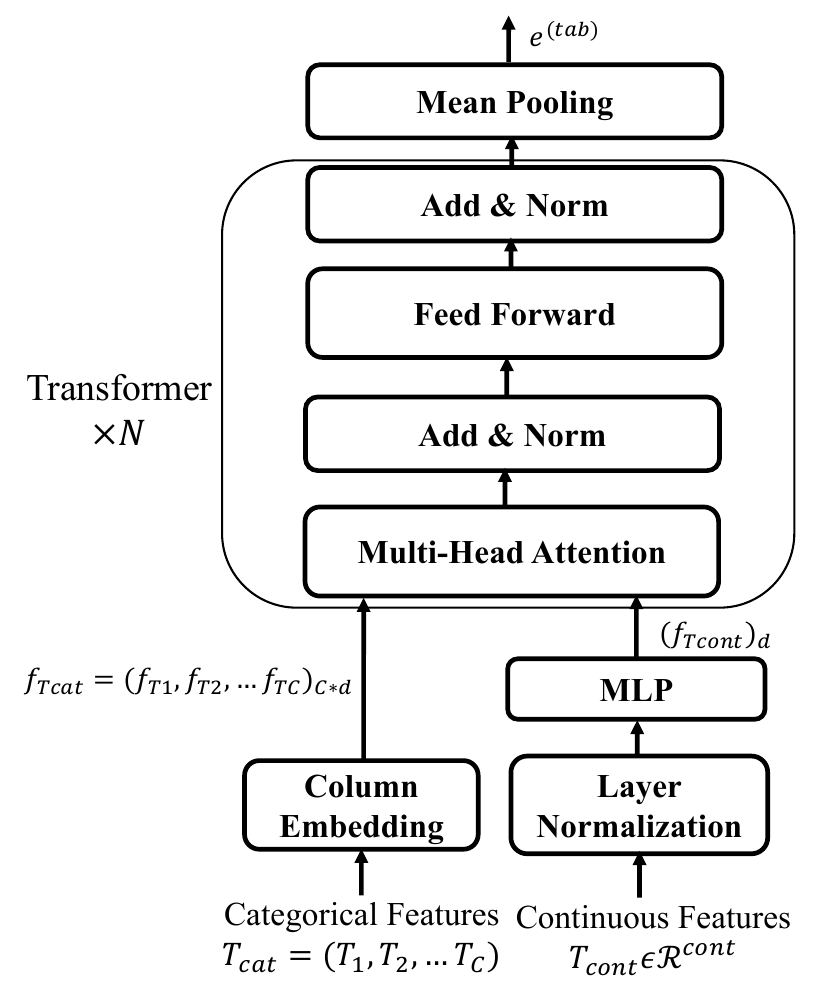}
    \caption{The architecture of our tabular encoder.
    }
    \label{fig:tab}
\end{figure}

For \textbf{Tabular Pre-training}, the architecture of the transformer-based encoder is shown in Fig. \ref{fig:tab}, as the categorical features $T_{cat}$ are encoded with Column Embedding\cite{tabtransformer} and continuous features $T_{cont}$ are encoded with MLP,
which are both fed into the transformer blocks. 
Followed by mean pooling operation among all features, tabular embedding $e^\text{(tab)}$ for each user is generated. In our model, transformer block number $N$ is selected as 4 and feature dimension $d$ is 1024.
To achieve the unsupervised pre-training, MLM and RTD tasks are performed. 
For MLM task, given a tabular input with categorical features $T_{cat} = \{ T_1, T_2, \cdot \cdot \cdot T_\mathcal{C} \}$, we randomly select $k\%$ features from index 1 to $\mathcal{C}$ and mask them as missing. By minimizing the objective that tries to predict the original features of the masked features, from the embedding outputted from the top-layer Transformer, MLM is conducted.
Here the selection rate $k$ is 30.
For RTD task, the original feature is replaced by a random value of that feature and the loss is designed to minimize for a binary classifier that tries to predict whether or not the feature has been replaced.
It can be noticed that different binary classifiers are defined for each column rather than a shared one, as each column has its own Column Embedding.
In our tabular pre-training process, $\lambda_{MLM}$ is set at 0.6 and $\lambda_{MLM}$ is 0.4.
The tabular pre-training takes $\sim$ 50 hours on 8 A100 GPUs for model convergence.

\subsection{User-Text Alignment}

After pre-training the user encoder, multi-modal features are fused and further aligned with text features generated from the future user behavior to form the user targeting system.

In multi-modal feature fusion, similar strategy based on cross-attention in previous works \cite{crossattn} is performed in our model, transferring from conventional image-text data modality to our behavior-tabular-text modal.
The dimension for each modal and the fused feature is 1024.
And for further user-text alignment, 
pre-trained multi-modal encoder followed with feature fusion module is the user encoder, while LLM with LoRA fine-tuning module is utilized as the text encoder.

The embedding of the user information and future text are generated from the previous-seven and next-two time windows, respectively. The time span is identical to User Behavioral Sequence Pre-training in \ref{self-supervise}, which means the user feature is generated by the multi-modal user data from the first 7 weeks and further aligned with text which is based on the template in \ref{user_pair} from the following 2 weeks using the contrastive learning.
The user-text alignment takes $\sim$ 70 hours on 16 A100 GPUs for model convergence.

\section{Additional Experimental Results}

\begin{figure}[!h]
    \centering
    \setlength{\abovecaptionskip}{0.1cm}
    \includegraphics[width=0.95\linewidth]{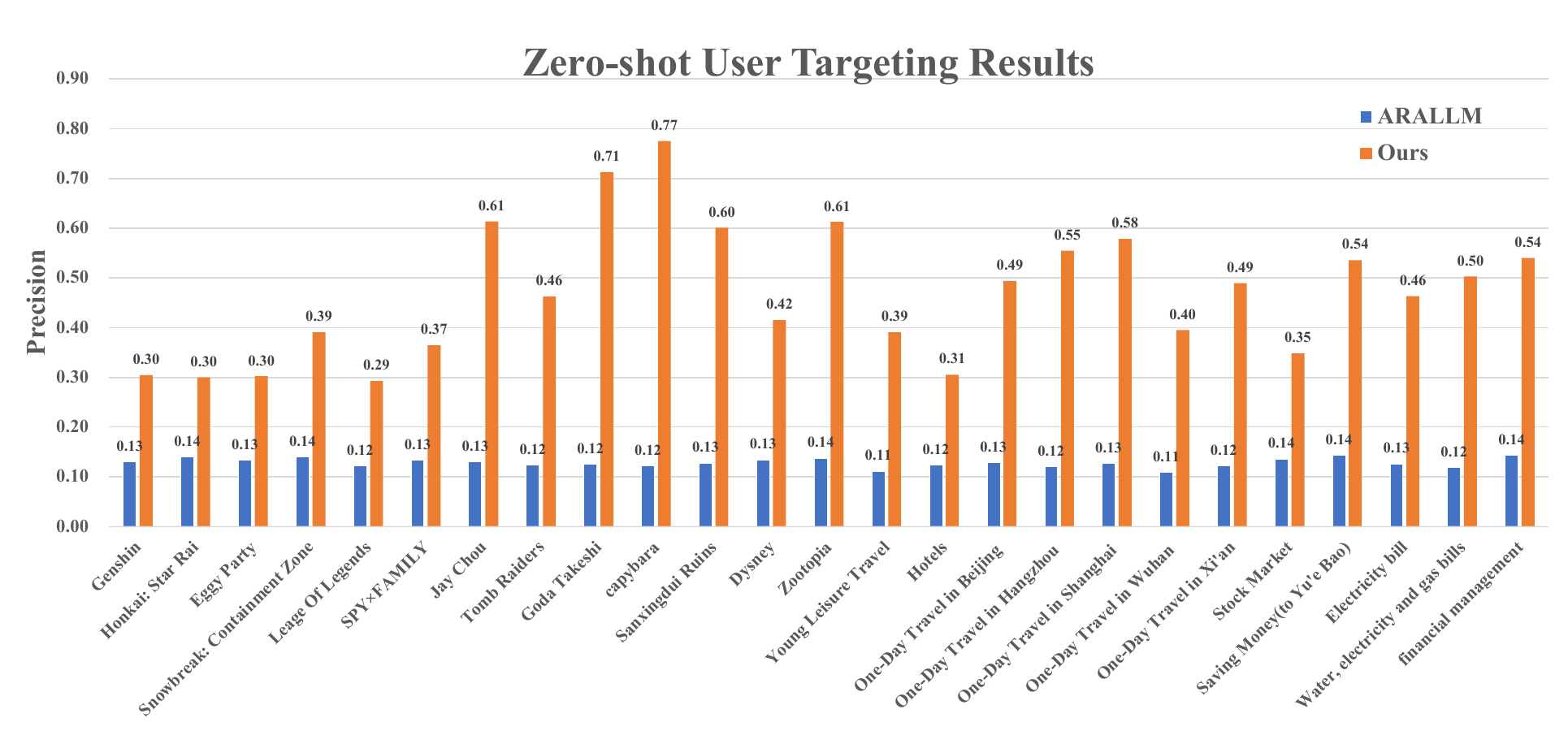}
    \caption{Zero-shot user targeting results in different real-world advertising application scenarios.
    }
    \label{fig:add_user_targeting}
\end{figure}

We present additional zero-shot user targeting results (measured by Precision on the validation dataset) across different real-world advertising application scenarios in Alipay. 
Fig \ref{fig:add_user_targeting} illustrates that our foundation model outperforms the previous model significantly.

\section{Limitations and Future Work}

Though achieving high performance in user targeting and user modeling, there are still some limitations in our model:

1. Though showing fantastic capability at Alipay platform, our model's generalizability to other platforms or industrial scenarios is not fully tested.

2. The training cost remains high though our two-stage strategy releases the requirement compared to the from-scratch strategy.

Therefore, our future work will focus on the following:

1. Enhancing the model transferability across platforms. We plan to achieve this by enhancing the model's ability for processing multi-modal user features (such as images, videos, knowledge graph, etc) and improving the model scalability for more data.

2. Reducing the time consumption for the foundation model pre-training.

\end{document}